\documentclass[conference]{IEEEtran}
\IEEEoverridecommandlockouts
\usepackage{cite}
\usepackage{amsmath,amssymb,amsfonts}
\usepackage{todonotes}
\usepackage{algorithmic}
\usepackage{graphicx}
\usepackage{textcomp}
\usepackage{comment}
\usepackage{xcolor}
\usepackage{url}
\def\BibTeX{{\rm B\kern-.05em{\sc i\kern-.025em b}\kern-.08em
    T\kern-.1667em\lower.7ex\hbox{E}\kern-.125emX}}

\newcommand{\linebreakand}{%
\begin{@IEEEauthorhalign}
\hfill\mbox{}\par
\mbox{}\end{@IEEEauthorhalign}}

\usepackage[left=0.75in, right=0.75in, top=0.75in, bottom=0.78in]{geometry}
\title{IdentiARAT: Toward Automated Identification of Individual ARAT Items from Wearable Sensors}

\author{
\IEEEauthorblockN{
Daniel Homm\textsuperscript{1,2}, 
Patrick Carqueville\textsuperscript{3}, 
Christian Eichhorn\textsuperscript{4}, 
Thomas Weikert\textsuperscript{1},
}
\IEEEauthorblockN{
Thomas Menard\textsuperscript{1},
David A. Plecher\textsuperscript{4},
Chris Awai Easthope\textsuperscript{1}
}

\IEEEauthorblockA{
\textsuperscript{1}\textit{Data Analytics \& Rehabilitation Technology (DART), Lake Lucerne Institute, Switzerland} \\
\textsuperscript{2}\textit{TUM School of Computation, Information and Technology, Technical University of Munich, Germany} \\
\textsuperscript{3}\textit{Professorship of Sport Equipment and Sport Materials, Technical University of Munich, Germany} \\
\textsuperscript{4}\textit{Research Group Augmented Reality, Technical University of Munich, Germany} \\
Correspondence to: chris.awai@llui.org
}
}

\begin{document}

\maketitle

\IEEEpubid{\begin{minipage}{\textwidth}\ \\[12pt] \centering
  © 2025 IEEE. Personal use of this material is permitted. Permission from IEEE must be obtained for all other uses, in any current or future media, including reprinting/republishing this material for advertising or promotional purposes, creating new collective works, for resale or redistribution to servers or lists, or reuse of any copyrighted component of this work in other works.
\end{minipage}}
\IEEEpubidadjcol

\begin{abstract}
This study explores the potential of using wrist-worn inertial sensors to automate the labeling of ARAT (Action Research Arm Test) items. While the ARAT is commonly used to assess upper limb motor function, its limitations include subjectivity and time consumption of clinical staff. By using IMU (Inertial Measurement Unit) sensors and MiniROCKET as a time series classification technique, this investigation aims to classify ARAT items based on sensor recordings. We test common preprocessing strategies to efficiently leverage included information in the data. Afterward, we use the best preprocessing to improve the classification. The dataset includes recordings of 45 participants performing various ARAT items. Results show that MiniROCKET offers a fast and reliable approach for classifying ARAT domains, although challenges remain in distinguishing between individual resembling items. Future work may involve improving classification through more advanced machine-learning models and data enhancements.
\end{abstract}


\section{Introduction}
\begin{figure}[t]
    \centering
    \includegraphics[width=\linewidth]{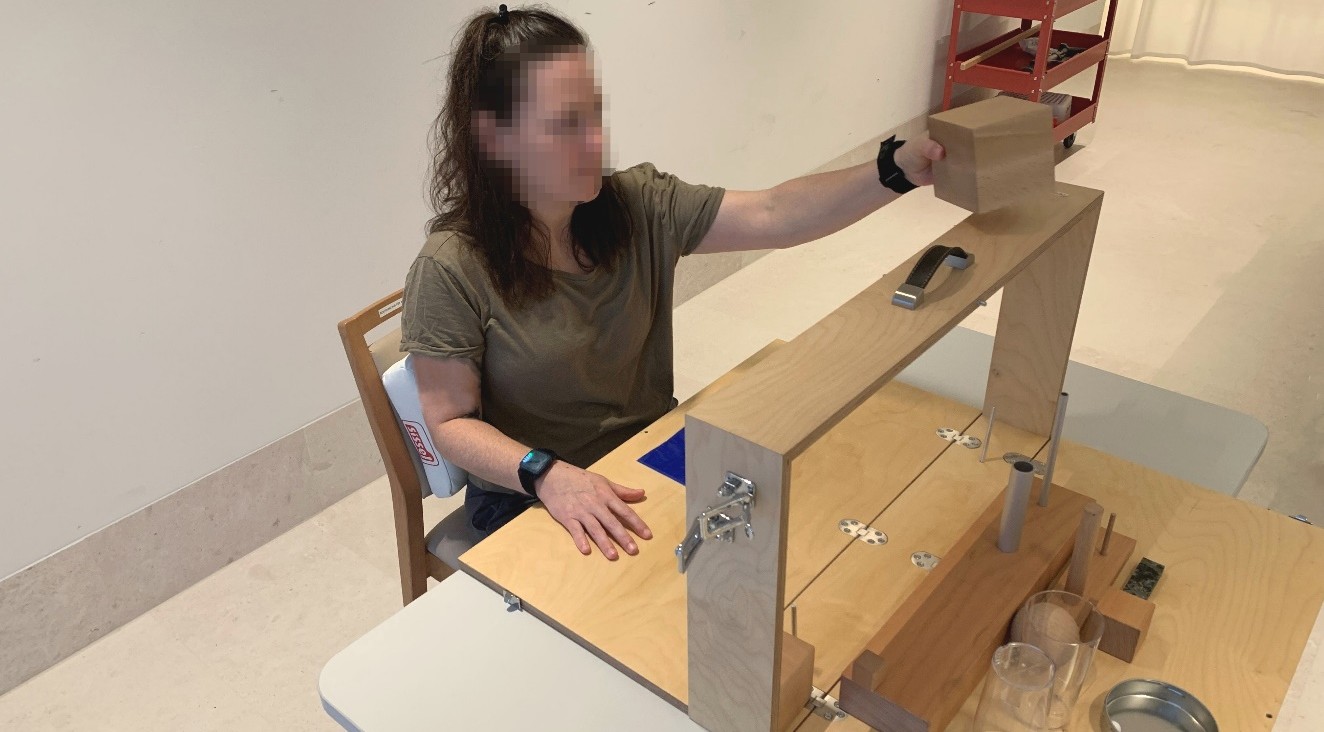}
    \caption{Patient performing ARAT item, Cube 10cm}
    \label{fig:cub-item-performed}
\end{figure}

Over 100 million people worldwide are living with the consequences of a stroke \cite{stroke-facts}, a condition that frequently results in long-term motor impairments. Of stroke survivors, 85\% present with initial upper limb impairments \cite{Thera-Intervention-PosStroke} and 60\% of survivors report motor disabilities that persist even after rehabilitation \cite{stroke-facts-initial}. Upper limb impairments span different domains and can include muscle weakness, changes in muscle tone, impaired motor control, and laxity of joints \cite{Review-IMU-Stroke-Upper-Limb}. Importantly, these upper limb impairments lead to reductions in motor function, which consequently have a negative impact on quality of life. Clinically, standard assessments exist to evaluate both impairment and function. These are important both to track rehabilitation progress, as also to evaluate intervention efficacy and optimize therapy content.
\IEEEpubidadjcol
The Action Research Arm Test (ARAT) is a popular standard clinical assessment for motor function that evaluates the capacity to perform functional movements. The test consists of 19 movements (items) of different complexities for each arm (e.g. Fig. \ref{fig:cub-item-performed}), which are sequentially performed and split into 4 subcategories (grasp, grip, pinch, gross). Each item is rated by an assessor on a scale of 0-3, based on the success, speed, and closeness to physiological motion. Scores are typically summed into subcategories to create a global score. The ARAT demonstrates good clinical properties with high interrater reliability, repeatability, and construct validity \cite{StandardizedARAT}. There is an ongoing discussion in the field as to what extent motor impairment may be approximated from the ARAT ~\cite{fugl-meyer1975}. 

Given that the ARAT is well established in clinical routine and shows promise to measure both impairment and function, there have been many approaches to investigate and better understand movement kinematics or at least statistical features of motion during the different movement items \cite{QQReachingMovements, ReviewKinematicAssessULMovements, StatisictsULReaching}. Instrumentation of the ARAT can give more details on the underlying kinematics, but at the same time, opens an opportunity for automation of the testing procedure. Automation is a necessary step for both clinical decision support and self-assessment, and a pre-requisite for any extension of ARAT measurements to the home. All of these topics are central pillars if rehabilitation as a field is to expand from a point-of-care to a continuum-of-care approach.

While motion capture systems and video recordings are well-suited to recognize human movements \cite{motion-capture-review}, these systems are expensive and can only be used in restricted environments. In contrast, wearable sensors such as inertial measurement units (IMUs) have no infrastructural requirements, are low-cost, and easy to use. These characteristics enable the use of IMUs at scale in a broad range of situations \cite{Survey_HAR_IMU}, including rehabilitation. Recent studies have demonstrated that IMU recordings are suitable for capturing essential characteristics of human movement across various tasks and activities \cite{Seenath2023, Young2022, Survey_HAR_IMU, IMU-STROKE-HOME-Kinematics}. Specifically, for upper limb movement assessments, IMU-based systems coupled with suitable analytics have been shown to deliver reliable and accurate results for many relevant movement metrics \cite{Review-IMU-Stroke-Upper-Limb}.

In terms of automation, previous work of our group has shown the ability to use wearable sensors to predict global ARAT assessor scores, achieving close to 80\% accuracy over all domains \cite{IMU-Stroke-Clinical-Scores-ARAT}. In the next step, we investigate whether wearable sensor data can be used to automatically identify each of the 19 individual ARAT items from wearable sensor signals, specifically triaxial acceleration and gyroscope data \clearpage in various different neurological populations. When working reliably, this automatic scoring proposal could be used as a clinical decision support system. For this purpose, we employ MiniROCKET \cite{MiniRocket}, a method that can efficiently be executed in real-time on wearable devices, ensuring feasibility for practical applications.

This objective gives rise to two research questions:
\begin{itemize}
    \item[] \textbf{RQ 1:} With which mean accuracy are we able to automatically identify individual items of the ARAT?
    \item[] \textbf{RQ 2:} To which extent does the impairment level affect the accuracy of the classification?
\end{itemize}
    
\section{Method}
\subsection{Study design}
We present a retrospective analysis (EKNZ 2024-00196) of instrumented ARAT data that was collected as part of the clinical routine at the cereneo - Center for Neurology and Rehabilitation from 1.1.2020 - 30.06.2024. All patients gave written informed general consent, which allowed their data to be used for retrospective analysis (EKNZ 2017-00730). 

\subsection{Data}
The ARAT belongs to part of the core clinical assessment battery of this participating clinic and was routinely performed by trained and experienced physiotherapists on in-patients, often at multiple time points. 
In this analysis, we included only patients with a present upper limb impairment, represented by Fugl-Meyer scores between 12-59. In total, this led to a dataset containing  2,725 correctly performed and labeled ARAT items. Additionally, items not contributing to ARAT scores due to early starts or supervision uncertainties were relabeled as \textit{junk}, adding 647 time sequences. However, this exceeded the number of each other label. To prevent the model from overfitting the junk data, we randomly removed junk sequences such that the number of junk labels was the same as that of the ARAT item with the most recordings. Of all the included individuals, 35 were men with a median age of 65 and 10 women with a median age of 70 years. The median height of all participants was 175 cm and the median weight was 55.2 kg. Five wearable sensors (Movella Dot) were mounted on each patient (Sensor location: left and right wrist, left and right middle of the upper arm, middle of the sternum), however, for all analyses presented here, only the two wrist sensors were used. These sensors were by far the best accepted and show the greatest promise for generalization to a home use case. The sensors recorded the following signals at a rate of 60~Hz: Triaxial acceleration [$\frac{m}{s^2}$], triaxial rotational acceleration [$\frac{deg}{s}$], and magnetic field strength [\textit{Gs}]. Magnetometer data was discarded, as deemed too unreliable due to a measurement location close to a clinical MRI. Orientations were calculated as quaternions on board the device using the VQF (Versatile Quaternion-based Filter) magnetometer-free estimation \cite{VQF}. Table \ref{tab:experiments-data-setup} shows all further preprocessing steps that were tested.

\subsection{Analytics}
\begin{figure}[t]
    \centering
    \includegraphics[width=0.5\textwidth]{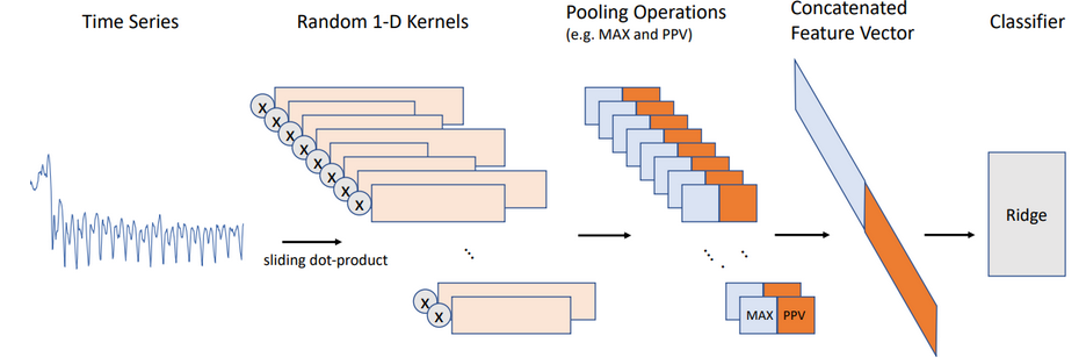}
    \caption{RandOm Convolutional KErnel Transform (ROCKET) approach \cite{Rocket-Figure}.}
    \label{fig:ROCKET-approach}
\end{figure}

To distinguish and correctly classify the labels of each time sequence, the ROCKET \cite{rocket} approach was used, as it is lightweight and well-suited to time series classification tasks. Specifically, we chose the MiniROCKET \cite{MiniRocket} adaptation as it is highly efficient, robust, and delivers comparable performance across diverse datasets \cite{rocket}. Table \ref{tab:MiniROCKET&rocket} highlights the differences from the original ROCKET method. MiniROCKET has demonstrated the ability to effectively capture patterns in Human Activity Recognition time series data \cite{MiniRocket, minirocket-for-HAR}. As shown in Fig. \ref{fig:ROCKET-approach}, the method first applies 1-D convolutional kernels and extracts feature values with pooling operations. The resulting feature vector is subsequently used as input for a classifier.

\begin{table}[t]
\caption{Setup differences ROCKET and MiniROCKET \cite{MiniRocket}.}
\resizebox{0.5\textwidth}{!}{
    \begin{tabular}{|c|c|c|}
    \hline
                       & ROCKET               & MiniROCKET                 \\ \hline
    length             & $\{7, 9, 11\}$       & 9                          \\ \hline
    weights            & $\mathcal{N}(0, 1)$  & \{-1, 2\}                \\ \hline
    bias               & $\mathcal{U}$(-1, 1) & from convolution output     \\ \hline
    dilation           & random               & fixed (rel. to input length) \\ \hline
    padding            & random               & fixed                      \\ \hline
    features           & PPV + max            & PPV                        \\ \hline
    number of features & 20,000               & 10,000                     \\ \hline
    \end{tabular}
}
\label{tab:MiniROCKET&rocket}
\end{table}

Table \ref{tab:MiniROCKET&rocket} shows the differences in the setup for ROCKET \cite{rocket} and MiniROCKET \cite{MiniRocket}. 
The table illustrates that ROCKET kernels have random lengths from the set $\{7, 9, 11\}$. Each kernel is initialized with weights $\sim \mathcal{N}(0, 1)$, bias terms $\sim \mathcal{U}(-1, 1)$, random dilations, and random paddings. For each applied kernel, the proportion of predicted positives (PPV) and the maximum value are calculated. These result in a default total of 20,000 features \cite{MiniRocket}. 

With MiniROCKET \cite{MiniRocket}, the randomness from ROCKET \cite{rocket} is partially removed by using kernels with a fixed length of 9. Furthermore, weights are chosen from the set $\{-1, 2\}$. The authors show that kernels should be as small as possible to maximize computational efficiency. $2^9~=~512$ possible kernels can be chosen for a kernel length of 9 and two possible weights. In \cite{MiniRocket}, an optimum of 84 is proposed to balance accuracy and computational cost. 
For a time series \textit{X}, kernel \textit{$W_d$}, and bias \textit{b}, the PPV is given by Equation \ref{PPV} \cite{MiniRocket}.
\begin{equation}
    PPV(X*W_d-b) = \frac{1}{N}\sum[(X*W_d-b) > 0]
    \label{PPV}
\end{equation}
For a kernel \textit{$W_d$}, dilation spreads the kernel over the input by applying it to every dth element, where \textit{d} is the dilation factor. In MiniROCKET, dilations are assigned from a fixed set based on the ratio of the input length and kernel length \cite{MiniRocket}.
The bias values \textit{b} are selected from the [0.25, 0.5, 0.75] quantiles of the convolution output for a given kernel \textit{W} and dilation \textit{d} on a randomly chosen training example. While this selection is random, the bias values themselves are derived from a fully deterministic convolution process, making MiniROCKET minimally random. Here, \textit{N} refers to the number of elements in the transformed time series for one sample. The only randomness in MiniROCKET is the selection of training examples for sampling bias values.

The advantage of PPV is that this feature summarizes key statistical properties of the time series, which often contain sufficient discriminatory information for classification tasks. In addition, while more features can provide more information, they can also introduce noise and overfitting to the model. Only using PPV reduces the risk of overfitting, especially on small datasets, as it acts as a regularizer by focusing on a single, robust feature.
In the end, MiniROCKET achieves almost the same accuracy as Rocket
using a mostly deterministic and much faster procedure \cite{MiniRocket}. For these reasons, MiniROCKET was used for our further analysis \cite{MiniRocket}.

\section{Experiments}

\begin{table}[]
\caption{The tested data preprocessing steps.}
\resizebox{0.49\textwidth}{!}{%
\begin{tabular}{|l|l|}
\hline
\textbf{Preprocessing}         & \textbf{Values}                \\ \hline
padding                        & Mean-padding; Zero-padding                     \\ \hline
Butterworth-filter frequencies & 10Hz; 20Hz; No                         \\ \hline
freed acceleration             & Yes; No                        \\ \hline
freed angular velocity         & Yes; No                        \\ \hline
gaussian noise for acceleration & $\mu$: 0 and $\sigma$: [0.1, 1]; no                         \\ \hline
gaussian noise for angular velocity & $\mu$: 0 and $\sigma$: [1, 5]; no                         \\ \hline
\end{tabular}
}
\label{tab:experiments-data-setup}
\end{table}

For the experiments, the open-source library sktime \cite{sktime} offers a built-in multivariate MiniROCKET implementation that was used for the classification tasks in this investigation.

\subsection{Validation}
We used a K-Fold Cross-validation (KFCV) with ten folds for all the experiments. This means we split the data into ten non-overlapping subsets (folds). Afterward, ten different models are trained on nine of the ten folds. For each model, the remaining tenth fold is subsequently used for validation. Here, the validation fold changes for each model until all folds are used exactly once for validation.
Standard KFCV is not ideal for time series data where temporal ordering matters between multiple sequences. However, in this research, the data set consists of independent time sequences. Therefore, we can split the data into folds, including complete item executions. 
Before splitting the data into the folds, the time sequences were shuffled individually with a random seed of 42 to retain reproducibility.
\subsection{Data Preprocessing}
In the first experiment, we took all 19 ARAT items and junk as individual classes. 
To determine which data preprocessing method works best, we tested all the combinations from Table \ref{tab:experiments-data-setup} on the data for the right and left wrists separately. Afterward, we chose the setup with the best mean accuracy for further investigation. 
During this process, we also observed the differences in the final accuracy. The model trained and tested on data from both wrists demonstrated the lowest accuracy. To provide an initial benchmark for potential performance on one wrist, we selected the wrist with the less accurate results as the focus for further investigation in this study.
\subsection{Input Features}
Having defined preprocessing and source data, we investigated which sensor recordings capture meaningful features for the classification task, therefore improving the model's performance and which input streams can be neglected to reduce computational costs. For this, the following three different sets of input streams were compared.
\begin{itemize}
\item 3-axis acceleration + 3-axis angular velocity
\item 4-axis quaternions
\item 3-axis acceleration + 3-axis angular velocity \newline + 4-axis quaternions
\end{itemize}
\subsection{Domain-Level Classification}

Each ARAT item belongs to a specific movement domain, \textit{Grasp, Grip, Gross,} and \textit{Pinch}. Therefore, with the best-performing setup for all individual items, we also investigated the classification possibilities of the different ARAT domains. This classification assessment helps evaluate whether patterns from individual items can be generalized across the domains, offering a more comprehensive understanding of the model's capabilities.

\subsection{Sequence Length Impact}
Finally, due to the large variation in data segment lengths, we investigated to what extent especially long sequences from the distribution tail impact the performance of the classification. For this, we trained one model on only 75\% of the shortest sequences.

\section{Results}
The accuracies presented in this section refer to the average results of all ten models from the cross-validation.
Due to the simplicity of MiniROCKET, calculating all embeddings and training a single model takes, on average, under 170 seconds on an Intel Core i5, 8th generation CPU, and 16GB RAM.
\subsection{Data Preprocessing}
The grid search for the best preprocessing resulted in the setups with the five best accuracies illustrated in Table \ref{tab:best-setups}.
\begin{table*}[t]
\caption{The five best data preprocessing setups and their mean accuracy results averaged for both arms. The data was reduced to 75\% of the shortest sequences to reduce time consumption.}
\resizebox{\textwidth}{!}{%
\begin{tabular}{|l|l|l|l|l|l|}
\hline
\textbf{Std. noise acceleration} & \textbf{Std. noise ang. vel.}& \textbf{Padding} & \textbf{Filter frequency acceleration $\backslash$ ang. vel.} & \textbf{Freed sensors} & \textbf{Mean accuracy} \\ \hline
0.5&3&Mean&0 / 0 & False & 64.5\% \\
0 & 0& Zero & 0 / 0 & False & 64.2\% \\
0.5 & 1 & Mean & 0 / 0 & False & 63.6\% \\
0.1 & 1 & Zero & 0 / 10 & False & 63.1\% \\
0.1 & 3 & Mean & 20 / 20 & False & 62.9\% \\ \hline
\end{tabular}
}
\label{tab:best-setups}
\end{table*}
The best result from Table \ref{tab:best-setups} includes a mean padding and an applied Gaussian noise with a mean of 0 and a standard deviation of 0.5 for the acceleration and 3 for angular velocity. Furthermore, the data was not freed from gravity, and no low-pass filter was applied. The preprocessing steps with the best result were applied before all other experiments.

We proceeded with the experiments with the data from the left wrist since the models trained and validated on this data split achieved, with around 5\% difference, less accurate results than models related to the right wrist data.
\subsection{Input Features}
\begin{figure}
    \centering
    \includegraphics[width=0.49\textwidth]{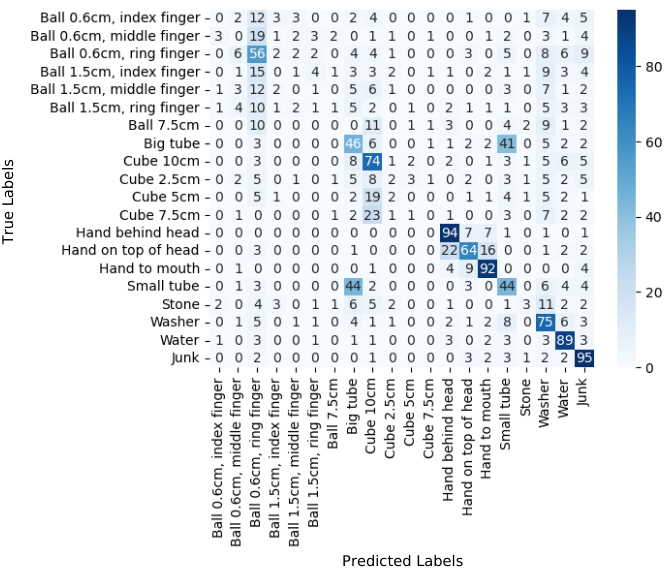}
    \caption{Confusion matrix of the aggregated validation results for the all-item classification, using the complete left wrist}
    \label{fig:all-labels_cm_all_data}
\end{figure}

The investigation of different input streams shows that including all available recording streams is beneficial. Using only acceleration and angular velocity as input features, the models were able to correctly classify about 42.7\% of the classes. In comparison, when working only with the calculated quaternions, the performance dropped to around 34.3\% accuracy. However, combining both sets of features improved the classification accuracy to 47.2\%. Fig. \ref{fig:all-labels_cm_all_data} shows the confusion matrix for the best-performing model with all ten input features for the left wrist.

\subsection{Domain-Level Classification}
\begin{figure}
    \centering
    \includegraphics[width=0.49\textwidth]{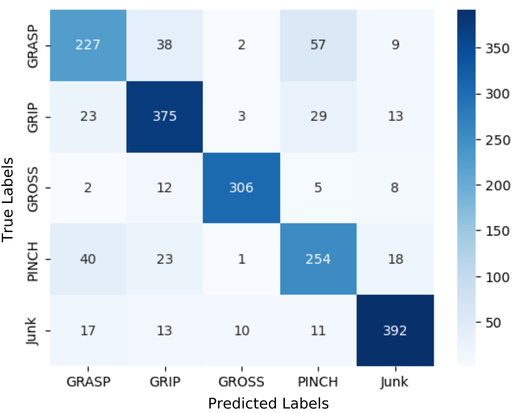}
    \caption{Confusion matrix of the aggregated validation results for the item-domain classification, using the complete left wrist data.}
    \label{fig:categorie-all-data}
\end{figure}

Even though these results do not seem promising for applied labeling of each item, Fig. \ref{fig:all-labels_cm_all_data} shows that most misclassifications belong to similar items. This leads to the result that for the classification of the domains, the same approach as for all 20 labels results in 82.3\% accuracy and the confusion matrix illustrated by Fig. \ref{fig:categorie-all-data}.

\subsection{Sequence Length Impact}
Discarding the 25\% longest sequences, as the final step of this experiment, led to two main insights.
The first main result concerns the effects of data sequence length. Removing the longest time sequences reduced the number of time sequences that achieved the  ARAT item scores zero or one from 16\% to only 3\%.
For the 937 left-over sequences, the count of each ARAT item was reduced by similar amounts. Removing the longest sequences reduced the maximal sequence length to $318 \pm 73$ time steps. 
Consequently, all paddings were reduced to around 10\% of the original length, reducing the run time by around 90\%. Removing the longest time sequences and therewith the most strongly impaired movements, the classification facilitates significantly. We observe an increase in performance to around 61\% accuracy for the classification of all ARAT items and to 93.9\% for the domain classification.

\section{Discussion}

The experiments of this investigation have demonstrated the possibilities of automatic ARAT item labeling using lightweight approaches.

\subsection{Validation}
In general, the KCFV approach allows us to use a large number of data for training. In addition, it provides a more stable and reliable estimation of the model's performance compared to a single train-test split by averaging the results of several train-test combinations. However, using an item-specific split leads to movement sequences of the same person being included in training and validation. While this can improve performance compared to the classification of a totally novel subject, it also favors models that overfit the training subjects.

\subsection{Experimental Results}
The first notable part of the preprocessing is that neither zero-padding nor mean-padding greatly impacts the resulting accuracy. This is because time series classification models like MiniROCKET primarily focus on patterns within observed data. MiniROCKET’s kernel-based feature extraction is robust to padding due to its invariance to constant offsets and reliance on PPVs, which are robust to zero padding. 
Secondly, filtering high frequencies from the data and freeing acceleration or angular velocity recordings from gravity result in lower accuracies. One underlying reason for this may be that the frequency filter and the freed inputs reduces the information from essential features. 
The last observation concerning preprocessing is the possible model improvement through additional noise. Noise can be characterized by its statistical moments and the range of values it spans. It should relate to the data's range and statistical moments. Noise should not distort nor overwhelm the original data recordings, ensuring that the underlying patterns remain intact for accurate analysis. However, noise can result in a more generalizable model when applied correctly. This is because noise can introduce variability into the training data, preventing the model from overfitting the training samples \cite{NoiseInjection}.

With regard to the decision to use the left wrist data. The lower performance for left wrist data, compared to right wrist data, reflects its more diverse score distribution. This diversity represents a more balanced dataset and reduces the overfitting risks, supporting its use in this study. As scores were item- and side specific, there was no requirement to define an affected and unaffected side.

The experiments with different input streams demonstrate that while IMU readings and quaternions individually provide valuable activity recognition features, each has its limitations. However, when combined, the model's performance improves. We can assume that the input features complement each other. For future research, we would recommend integrating both inputs to enhance performance.

With regard to research question one (\textbf{RQ 1}), the final results indicate that while the current approach effectively disambiguates ARAT domains, fine-grained classification of individual items within a single domain remains limited. Addressing these challenges will require incorporating additional sensors, features, or enhancing feature extraction techniques in future work.

Fig. \ref{fig:categorie-all-data} demonstrates that the ARAT domains can be effectively distinguished with the current dataset using a lightweight and straightforward approach. The most frequently confused classes by the model are \textit{Grasp} and \textit{Pinch}. This is likely due to challenges in differentiating between certain \textit{cube} items within the \textit{Grasp} domain and \textit{ball} items within the \textit{Pinch} domain using only a wrist mounted IMU which encodes no information on the finger motion.

In terms of clinical utility, the current classification accuracy can only be considered as a proof of concept. The use of a minimal wearable sensor set reduces set up time and complexity, two very relevant aspects for clinical and home use. The addition of more wearable sensors will reduce feasibility. However, sensor fusion between video systems and wearable sensors may provide a straightforward solution to balancing data-greediness with practicality. 

Regarding the second research question (\textbf{RQ 2}), the last experiment highlights the retaining large variance in the length of item sequences (ie. item duration) significantly complicates the classification task. Long item sequences were mainly caused by inaccurate use of the start/stop function, which resulted in some sequences being 10x longer than the ARAT cutoff times. Slow executions are also frequently linked to greater impairment, hence our decision to remove the slowest 25\% disproportionately affected scores 0 and 1, of which the relative proportion dropped drastically. This makes it very challenging to disentangle the confounding effects of decreased execution velocity and reduced movement quality. Adding well-understood hand-crafted metrics such as smoothness and trunk compensation metrics to the model may support do disentangle velocity effects through improved inference. None-the-less, reducing temporal variation of item length significantly boosted prediction accuracy for the current approach. This also would be a further incentive for examining ARAT item time cut-offs at an item-specific level, as opposed to the current "one-time-fits-all" approach.

\subsection{Limitations and Future Work}
In future work, we would aim to better disentangle execution velocity and compensation from movement quality. Furthermore, we would endeavor to develop multi-stage classification models to better handle class 0 \& 1 scores, or add patient-specific priors that generalize across tasks. Finally, we would add additional sensing modalities to the data collection to be able to incorporate finger motion into our classification. This will allow us to improve the fine-grained item classification within single domains.  
From a model perspective, we would consider more data hungry approaches such as transformers which we would pre-train on public datasets or with synthetic data.

\section{Conclusion}

We developed an auto-labeling model for individual ARAT items using two IMUs on the wrists. With its fast runtime and simplicity, MiniROCKET functions as a practical proof of concept to automatically classify ARAT items in neurological patients. The findings demonstrate that meaningful features can be effectively extracted from wrist-worn sensors. Despite the item classification still needing improvement, MiniROCKET allows for reliable differentiation between task domains. Finally, the analysis demonstrates that longer and more variable item sequences, typically associated with greater impairment levels, significantly decrease classification accuracy.

\section{Acknowledgments}
We thank our clinicians: Alma Flueck, Migjen Shala, Elena Ruiz, Lisa Herndlhofer, our student team, and all participants.

\bibliographystyle{ieeetr}
\bibliography{ref}
\vspace{12pt}
\end{document}